\documentclass[akbc,twoside,11pt,lettersize]{article}
\usepackage{akbc}
\usepackage{graphicx}
\usepackage{caption}

\newcommand{\eat}[1]{}

\usepackage{quoting}
\newenvironment{myquote}{                   
  \parskip 1mm \begin{quoting}[vskip=0mm,leftmargin=10mm]}{
\end{quoting}}
\newenvironment{ite}{                     
  \vspace{-2mm} \parskip 0cm \begin{itemize} \parskip 0cm \parsep 0cm \itemsep 0cm \topsep 0cm}{
	\end{itemize} \vspace{-2mm} }
\newenvironment{enu}{                   
     \parskip 0cm \begin{list}{}{\parsep 0cm \itemsep 0cm \topsep 0cm}}{
       \end{list}} 
\newenvironment{des}{                 
     \parskip 0cm \begin{list}{}{\parsep 0cm \itemsep 0cm \topsep 0cm}}{
       \end{list}} 

\newcommand{\kburl}{https://allenai.org/data/haspartkb}
\newcommand{\genkburl}{https://allenai.org/data/genericskb}

\ShortHeadings{A New Knowledge Base of hasPart Relations}{A New Knowledge Base of hasPart Relations}
\finalcopy 

\begin{document}

\title{Do Dogs have Whiskers? \\A New Knowledge Base of hasPart Relations}

\author{\name Sumithra Bhakthavatsalam \email sumithrab@allenai.org \\
\name Kyle Richardson \email kyler@allenai.org \\
    \name Niket Tandon \email nikett@allenai.org \\
       \name Peter Clark \email peterc@allenai.org \\
       \addr Allen Institute for AI, Seattle, USA}

\maketitle
\begin{abstract}
We present a new knowledge-base (KB) of hasPart relationships, extracted from a large
corpus of generic statements. Complementary to other resources available, it is the
first which is all three of: accurate (90\% precision), salient (covers relationships
a person may mention), and has high coverage of common terms (approximated
as within a 10 year old's vocabulary), as well as having several times more
hasPart entries than in the popular ontologies ConceptNet and WordNet.
In addition, it contains information about quantifiers, argument modifiers,
and links the entities to appropriate concepts in Wikipedia and WordNet.
The KB is available at \kburl
\end{abstract}

\section{Introduction}
\vspace{-2mm}
\label{Introduction}

Meronymic (hasPart) relations are one of the most important and frequently
used relationships in reasoning systems, perhaps second only to the
generalization (``isa'') relationship. hasPart knowledge plays a role in
multiple inference scenarios, for example:
\begin{ite}
\item If X moves to Y, then X's parts move to Y;
\item If part of X is broken, then X is broken;
\item To construct X, one needs all the parts of X;
\item If X is ill, then some part of X may be the cause;
\end{ite}
However, while there has been extensive research on mining hasPart relationships
from text, e.g., \cite{girju2006automatic,Hage2006AMF,Ling2013ExtractingMF}, 
there are only a few resources that
have been made available. Two popular
general resources, WordNet \cite{fellbaum1998wordnet} and ConceptNet \cite{speer2017conceptnet},
contain collections of only 9k and 13k hasPart relationships respectively.
In addition, when restricted to hasPart relations between common terms,
which we approximate as within the typical vocabulary of a Fifth Grader (age 10) \cite{stuart2003children},
these totals drop to $\approx$1k in each resource.
Other resources have different limitations: Quasimodo \cite{Romero2019CommonsensePF} contains 18k partonomic
relationships, but only covering body parts rather than the general hasPart
relationship; WebChild \cite{tandon2014webchild} contains 256k hasPart relations, but a large proportion
covers unusual concepts - only 9k are within a Fifth Grade vocabulary;
and although the resource PWKB (part-whole KB) \cite{Tandon2016CommonsenseIP} contains 6.5M hasPart
relations, the large majority were computed by an inference-based expansion of a smaller set,
resulting in many entries that a person would be unlikely to mention (low salience).

We contribute a complementary resource of hasPart knowledge,  the 
first which is all three of: accurate (90\% precision), salient (covers relationships
a person may mention), and has high coverage of common terms (within a Fifth Grade vocabulary).
While our main contribution is the resource itself,
our approach to extraction is also novel: rather than extracting hasPart
relations from arbitrary text, we only extract from {\it generic sentences},
i.e., statements about members of a category such as ``Dogs have tails.''.
Empirically, this results in a high yield of good quality extractions (Section~\ref{evaluation}),
and significantly higher than a strong, prior extraction pipeline applied to the same corpus (Section~\ref{tuplekb-comparison}).
Our resulting hasPartKB contains over 50k entries,
including over 15k within a high-schooler's vocabulary, each
additionally annotated with information about quantifiers, argument modifiers,
and links the entities to appropriate concepts in Wikipedia and WordNet.

Our work is targeted in two important ways. First, although there are several types
of hasPart relationship (e.g., Winston et al. (\citeyear{winston1987taxonomy}) provide a
commonly used taxonomy of six types,\footnote{
  component/integrated object, member/collection, portion/mass, stuff/object,
  feature/activity, and place/area.}; similarly Keet \& Artale (\citeyear{Keet2008RepresentingAR})
axiomatize a taxonomy of ten types), their mention in language is highly skewed
towards (from Winston et al.'s taxonomy) two types, namely ``component/integral object'' (e.g., a handle is part of a cup)
and ``stuff/object'' (e.g., steel is part of a bike). As a result,
we bound the scope of our work to just these two types. Second
we target {\it salient} parts, which we informally define as
those that a person might consider mentioning, and by implication
more likely to be useful in an end-task. The restriction is important
as, in a literal sense, many entities have millions of parts, making
a complete enumeration both infeasible and unhelpful.\footnote{
  For example, WordNet's 9k hasPart relations expand to a database
  of 5.3M parts when inheritance and transitivity is applied exhaustively,
  including entries such as ``A nucleolar organiser is part of a poet Laureate''
  and ``A bedspring is part of a dude ranch''.}

\vspace{-2mm}
\section{Related Work}
\vspace{-2mm}
There has been substantial prior work on extracting hasPart knowledge,
although the majority did not result in publically available resources
being released. Early work by Berland and Charniak (\citeyear{berland1999finding}) used
two Hearst patterns \cite{hearst1992automatic} to extract part-wholes,
but just covering a small number of objects. Girju et al. \cite{girju2006automatic,girju2003learning}
developed a semi-automatic method for extracting part-whole relations, using
a combination of hand-identified lexical patterns and machine-learned
selectional constraints to optimally filter extractions, resulting
in 10k extractions at 80\% precision (although no public resource was released).
Similarly, Ling et al (\citeyear{Ling2013ExtractingMF}) combined distant supervision
with multi-instance learning for meronymy extraction,
in particular aggregating evidence from multiple sentences together
to reduce noise. However, the method was only applied to biology
text and again no resource was released. More recently,
Tandon et al. developed WebChild (\citeyear{tandon2014webchild}), a large-scale resource including
hasPart relations, and PWKB (part-whole KB) (\citeyear{Tandon2016CommonsenseIP}) specifically
aimed at hasPart relations. WebChild includes 256k hasPart relations (comprising a subset of most reliable PWKB relations), while PWKB contains 337k core relations, expanded to 6.5M entries using rules for
inheritance and transitivity - we compare our results to WebChild and PWKB in
Section~\ref{evaluation}. PWKB's relations were extracted by first
finding part-whole lexical patterns (e.g., ``{\it noun} of the {\it noun}'') using 1200
seed part-whole pairs from WordNet, applying those seed patterns to a large corpus with
a novel scoring function, and finally expanding and filtering the results using
inheritance and transitivity rules. Seed patterns have similarly been used to
bootstrap extraction elsewhere, e.g., \cite{Pantel2006EspressoLG,Ittoo2013MinimallysupervisedEO}.
Two other related, general resources are
TupleKB \cite{tuplekb} and Quasimodo \cite{Romero2019CommonsensePF},
both general KBs including hasPart knowledge. However, although TupleKB is large (280k entries),
it contains less than 1000 hasPart entries (excluding those added directly from WordNet),
so has limited hasPart coverage. Likewise, Quasimodo only includes the ``has body part''
relation in its extraction vocabulary, rather than general meronymic relationships.

In the last few years, general relation extraction has largely shifted to using neural
techniques, e.g., \cite{Lin2016NeuralRE,kuang2019improving,Wang2019ExtractingMI}.
We similarly apply neural methods (using BERT \cite{bert} and RoBERTa \cite{roberta}),
but specifically for hasPart, and do not claim any novelty
in this aspect of our approach.


\vspace{-2mm}  
  \section{Approach}
\vspace{-2mm}
Our approach to hasPart extraction has five steps:
\begin{enu}
\item[1.] Collect generic sentences from a large corpus
\item[2.] Train and apply a RoBERTa model to identify hasPart relations in those sentences
\item[3.] Normalize the entity names
\item[4.] Aggregate and filter the entries
\item[5.] Link the hasPart arguments to Wikipedia pages and WordNet senses
\end{enu}
We now describe each step in turn. 

\vspace{-2mm}
\subsection{Step 1: Collecting Generic Sentences $GKB$ \label{step1}}
\vspace{-2mm}
Rather than extract knowledge from arbitrary text, we extract hasPart relations
from {\it generic sentences}, e.g., ``Dogs have tails.'', in order to bias the process
towards extractions that are general (apply to most members of a category) and
salient (notable enough to write down). As a source of generic sentences,
we use GenericsKB, a large repository of 3.4M
standalone generics previously harvested from a Webcrawl of 1.7B sentences \cite{Bhakthavatsalam2020GenericsKBAK}.\footnote{
  GenericsKB is available at \genkburl}
GenericsKB was constructed by first using a set of lexico-syntactic rules 
to identify candidate standalone generic sentences, and then applying 
a crowdsource-trained BERT classifier to assign a confidence to
each generic. For our task here, we use only the highest-ranked
generics, namely sentences with an associated confidence of $>$ 0.5.
This subset contains 386k sentences, to use for the
subsequent hasPart extraction. We will refer to this subset as $GKB$ in the rest of this paper.

\vspace{-2mm}
\subsection{Step 2: hasPart Extraction}
\vspace{-2mm}
To identify hasPart relationships in a sentence $S$, we first identify candidates and then
train and apply a RoBERTa model to classify them, as we now describe.

First, for each sentence $S$ in $GKB$, we identify all noun chunks in the sentence using
a noun chunker (spaCy’s Doc.noun\_chunks). Each chunk is a candidate whole or part.
Then, for each possible pair, we use a RoBERTa model to classify whether a
hasPart relationship exists between them. The input sentence is presented to RoBERTa
as a sequence of word piece tokens, with the start and end of the candidate hasPart arguments
identified using special tokens, e.g.:
\begin{myquote}
  {\it [CLS] [ARG1-B]Some pond snails[ARG1-E] have [ARG2-B]gills[ARG2-E] to breathe in water.}
\end{myquote}
where {\it [ARG1/2-B/E]} are special tokens denoting the argument boundaries. The
{\it [CLS]} token is projected to two class labels (hasPart/notHasPart), and a softmax
layer is then applied, resulting in output probabilities for the class labels. We
train with cross-entropy loss. 
We use RoBERTa-large (24 layers), each with a hidden size of 1024, and 16 attention heads, and a total of 355M parameters.
We use the pre-trained weights available with the model and further fine-tune the model parameters by training on our labeled data for 15 epochs.

To train the model, we use a hand-annotated set of $\sim$2k examples. Given that sentences
expressing hasPart information
are sparse in $GKB$, collecting a representative sample of positive and negative examples to annotate
is itself a challenge. To help with this, we proceeded as follows:
\begin{enu}
  \item[1.] Train a similar model, distantly supervised using a subset of ConceptNet's partOf relations:\footnote{ConceptNet
  has $\sim$13k partOf relations which are noisy. We used a combination of heuristics and manual filtering
  on this set to reduce it down to a set of $\sim$9k more reliable partOf relations.}
$GKB$ sentences mentioning {\it both} terms in a ConceptNet partOf relation are used as positive
examples. Negative examples were generated by (a) reversing the arguments in positive
examples, and (b) using $GKB$ sentences mentioning arguments from {\it other} ConceptNet relations besides partOf.
The result is a coarse-grained hasPart classifier based on ConceptNet's data.
\item[2.] Apply this model to each $GKB$~sentence, for each pair of noun chunks it contains,
  to find sentences that likely contain a hasPart relation. We treat these as good candidates
  to hand-annotate. We take a sample (380) of such sentences for this purpose.
\item[3.] For {\it all} noun chunk pairs in each sample sentence, we annotate each to indicate if
  a hasPart relationship holds or not. This process resulted in a final training set of 2,106 training examples,
  used to train the final RoBERTa-hasPart-classifier model. 
\end{enu}

After training the model, we run it over all sentences $S$ in $GKB$, and for all noun chunk pairs in $S$,
to classify each pair as a hasPart relation. Of the (several million) classifications, 
we obtain a total of $\sim$127k hasPart examples (hasPart class score $>$ 0.5) in the
initial hasPart database. We now normalize, aggregate, filter, and link the
this data, described next.


\vspace{-2mm}
\subsection{Step 3: Entity Normalization}
\vspace{-2mm}
The noun chunker sometimes identifies chunks that include quantifiers (e.g., ``most'') and/or modifiers (e.g., ``large'').
To normalize the entity names, we remove these, but retain them as metadata.
Quantifiers are identified simply by checking if the first word of the entity name is a quantification word (using a small list of quantifier words).
Modifiers are identified by cross-referencing with Wikipedia titles: If the entire entity name is a Wikipedia title, it is retained.
If not, the first word is removed as a possible modifier, and the shortened name is again checked against Wikipedia. This process
is repeated until a Wikipedia name is found. If no Wikipedia name is found even after removing all words, the entire entity name is retained.
In this way, entity names  such as ``large elephant'' will become ``elephant'' modified by ``large'', while ``large intestine'' will
be retained as a single term. 

\vspace{-2mm}
\subsection{Step 4: Aggregation and Thresholding}
\vspace{-2mm}

We post-processed these to aggregate duplicate tuples resulting from multiple sentences.
The aggregated tuples are assigned an average-pooled score from the scores from individual sentences.
To further improve precision (though at the expense of recall), we removed hasPart relations whose
RoBERTa score was below a threshold (chosen to raise precision to 90\%). The resulting yields and precisions (measured by sampling) are shown in Table~\ref{filtering}.

\begin{table}
  \centering
  {\small
  \begin{tabular}{|l|l|c|c|} \hline
    {\bf Filter} & {\bf Yield} & {\bf Precision (\%)} \\ \hline
    None & 78,343 & 80 \\
    Score $\leq$0.9985 (score-cutoff) & 50,752 & 90 \\ \hline
\end{tabular}
}
  \caption{After thresholding, the final hasPart KB contains 50k entries, with precision of $\approx$90\% (based on sampling).}
  \label{filtering}
\end{table}

\vspace{-2mm}
\subsection{Step 5: Word Sense Disambiguation and Entity Linking \label{wsd}}
\vspace{-2mm}
Finally, the two entities in each hasPart entry are linked to both their WordNet word sense, and a Wikipedia page (using the Wikipedia title span identified in the earlier Step 3). 

For assigning word senses, we reimplemented the GlossBERT word sense disambiguation (WSD) model \cite{Huang2019GlossBERTBF}. GlossBERT takes the sentence + the word to disambiguate + and a gloss of a possible sense of the word, and outputs a score. Softmax'ing scores over all possible senses yields the most likely sense. In contrast to the original paper, however, we used the more recent RoBERTa encoder instead of BERT, and employed a modified ranking loss function \cite{Richardson2019WhatDM}. On existing WSD benchmarks, our implementation obtains results comparable to state-of-the-art (See Appendix). As a sanity check, we also manually sense-labeled a random sample of 70 entities (in context) in the hasPartKB. Of these, 41 were polysemous, and 38 of the 41 were assigned the labeled sense by our WSD model, suggesting high accuracy in sense assignment.

Entity linking (to a Wikipedia title) follows naturally from the earlier Step 3, where entity normalization included preferring names that are also
Wikipedia titles. This task is substantially less ambiguous than WordNet sense assignment, as Wikipedia titles are coarser grained and topical, so
ambiguity is rarer. If the Wikpedia title is ambiguous (links to a disambiguation page) we omit the link.
In our database, 87\% of the entities have associated titles (including disambiguation pages), and of these 76\% are unambiguous, hence linked (= 66\% linked overall). A more sophisticated approach would be to use the sentence context to diambiguate the title for ambiguous cases.

\vspace{-2mm}
\subsection{Final hasPartKB}
\vspace{-2mm}
The resulting hasPart database contains over 50k entries with a (sampled) precision of $\approx$90\%.
In addition, each entry contains information about quantifiers, argument modifiers,
and links the entities to appropriate concepts in Wikipedia and WordNet. 

\eat{
  \vspace{-2mm}
  \subsection{Taxonomy Building \label{taxonomy-builder}}
\vspace{-2mm}
In addition to the hasPart database, we explored constructing coherent hasPart taxonomies
for different entities. Simply traversing hasPart links based on entity names can cause errors to compound,
e.g., if there is a word sense change, or the hasPart link has lower confidence. Thus to
improve quality of the taxonomy, we used three additional constraints:
\begin{ite}
\item {\bf Topicality:} We only use hasPart entries where one of the arguments is the subject of the supporting sentence. For example, from ``Elephants are animals with trunks'' supports {\it hasPart(elephant,trunk)} but not {\it hasPart(animal,trunk)} (if the latter had mistakenly been extracted).
\item {\bf Quantification:} Building paths from relations that are explicitly quantified with an existential quantifier compounds uncertainty and errors. To mitigate this, we add a link to the path only if the new link is not explicitly quantified with one of \{some, few\}. 
\item {\bf Word Sense:} Expand paths in a sense-constrained manner, using the word sense tags assigned earlier. During taxonomy building, we add a link only if the new link refers to the same word sense for the shared argument.
\end{ite}
We evaluate the impact of these constraints shortly.
}
  
\vspace{-2mm}
\section{Evaluation \label{evaluation}}
\vspace{-2mm}
We now evaluate hasPartKB along three dimensions: precision, coverage, and salience,
and compare these to several existing resources with hasPart data: 
WordNet
\cite{fellbaum1998wordnet}, ConceptNet \cite{speer2017conceptnet}, Quasimodo 
\cite{Romero2019CommonsensePF}, 
TupleKB \cite{tuplekb}, WebChild \cite{tandon2014webchild}, and PWKB \cite{Tandon2016CommonsenseIP}. Note that these three metrics
interact, thus no single measure should be taken in isolation.

\vspace{-2mm}
\subsection{Precision}
\vspace{-2mm}
Table~\ref{precision} shows the (approximate) precision of our KB and other resources,
using either the published precision figures, or human judgements over a small (200)
sample of randomly selected entries. The main observation is that {\bf all
  the resources have good precision ($\approx$80\%+)}, reflecting the respective
care that has gone into their construction.

\begin{table}
  \centering
  {\small
  \begin{tabular}{|l|cc|} \hline
          {\bf Source} & \multicolumn{2}{|c|}{\bf Precision (\%)} \\ \hline
          Wordnet 3.0 & 100\% & (from \cite{Tandon2016CommonsenseIP}) \\
          ConceptNet 5.6.0 & 91\% & (sampled) \\
          Quasimodo 1.2  & 78\% & (sampled) \\
          TupleKB & 80\% & (from \cite{tuplekb}) \\
         Webchild & 82\% & (from \cite{tandon2014webchild}) \\
         PWKB & 89\% & (from \cite{Tandon2016CommonsenseIP}) \\
          hasPartKB (ours) & 90\% & (sampled) \\
          \hline
\end{tabular}
}
\caption{Precision of hasPart entries in the differing resources. All show good ($\approx$80\%+) precision for their entries.}
\label{precision}
\end{table}

\vspace{-2mm}
\subsection{Coverage}
\vspace{-2mm}
We now evaluate the coverage of our hasPartKB and the other resources.
We also conduct a small case study on coverage of six selected concepts,
using independently authored parts lists for each.

\vspace{-2mm}
\subsubsection{Database Sizes (Yield)}
\vspace{-2mm}
To what extent does the KB comprehensively tell us the
parts of entities? While the notion of coverage is hard to define, we use two
approximations: First, what is the overall yield (size) of the database, and
second, what is the yield (size) when constrained to ``common'' concepts, which
we approximate as those with names
within the vocabulary of a Fifth Grader. In addition, for the
Fifth Grade subset, we count how many distinct wholes and how many
distinct parts are mentioned. Although these
are approximate measures, they provide some indicators of coverage.

Table~\ref{yields} show the comparative sizes of the full hasPart databases,
and the subset within a Fifth Grade vocabulary. We observe that:
\begin{ite}
\item PWKB has the largest coverage. However, this is largely due to its
  inference-based construction process, resulting in most entries being
  obscure (low salience, Section~\ref{salience}).
\item Of the remainder, although WebChild has the largest general
  coverage, our hasPartKB has 50\% greater coverage of hasPart
  relations between common terms (Fifth Grade vocabulary).
  This suggests the hasPartKB has greater coverage of
  core relationships, while WebChild has a broader coverage
  of less common concepts.
\end{ite}

Table~\ref{part-whole-counts} shows the number of distinct terms in the Fifth Grade
Vocabulary subset. By this metric, the results show hasPartKB has the
greatest coverage of parts, and, apart from PWKB, also for wholes within this core vocabulary.

\eat{
hasPartKB has a total of 15128 distinct entities. Following is the argument-wise breakdown:
\newline
\indent Total no. of entities that have at least one part: 9921
\newline    
\indent Total no. of entities that are part of at least one other entity: 8085

We find that the most frequently occurring entities in our KB are from the science domain. Figure 1 shows a plot of the most frequently occurring entities as parts and as wholes.
To assess the coverage of elementary school science concepts, we used a 5th Grade Vocabulary\cite{stuart2003children} and computed recall against it. Table 3 presents the total yield of hasPart relations in our KB (hasPartKB) and the other resources. While Quasimodo has a much higher coverage of Science hasPart relations than Wordnet and Conceptnet, its knowledge is strictly about body parts.
}

\begin{table}
  \centering
  {\small
  \begin{tabular}{|l|c|c|} \hline
          {\bf Source} & {\bf All} & {\bf in 5th Grade Vocab} \\ \hline
          Wordnet 3.0 & 9098 & 1399 \\
          ConceptNet 5.6.0 & 12988 & 2164 \\
           Quasimodo 1.2  & 3197 & 1546 \\
           TupleKB & 898 & 645 \\
          WebChild & 256k & 10566 \\
          PWKB & 6.5M & 336k \\
          hasPartKB (ours) & 50752 & 15200 \\ \hline
\end{tabular}
}
\caption{hasPart Relation Yield. Our hasPartKB has greatest yield over common (5th Grade)
terms, with the exception of PWKB. However, PWKB suffers from low salience (Section~\ref{salience}).}
\label{yields}
\end{table}

\begin{table}
  \centering
  {\small
  \begin{tabular}{|l|c|c|} \hline
    {\bf KB} & {\bf \#wholes} & {\bf \#parts} \\ \hline
        Wordnet 3.0 & 787 &  1068 \\
        ConceptNet 5.6.0 & 1125 & 1497 \\
         Quasimodo 1.2  & 836 & 119 \\
         TupleKB & 109 & 58 \\
        Webchild &  2061 & 496 \\
        PWKB & 12179 & 995 \\
        hasPartKB (ours) & 3294 &  3304 \\ 
        \hline
        \end{tabular}
}
\caption{The number of distinct Wholes and Parts, for entries within a 5th Grade Vocab. \label{part-whole-counts}}
\end{table}

\vspace{-2mm}
\subsection{Salience \label{salience}}
\vspace{-2mm}
Many objects have thousands, or even millions, of parts (including electrons and quarks),
making a complete enumeration both infeasible and unhelpful. Rather, we wish
to collect hasPart relationships that are likely to be useful. We refer to
such parts as {\it salient} parts. As a rather approximate indicator
of salience, we consider an entry salient if it is one that ``someone might reasonably consider mentioning.''.
For example, ``a tail is part of a dog'' is salient, but ``a vacuole is part of a queen consort'' is not.
We can weakly operationalize this by, given {\it hasPart(x,y)}, checking whether
there is a sentence mentioning both $x$ and $y$ in a large corpus, i.e.,
the relationship has (likely) been mentioned.\footnote{
  Of course, the sentence may be describing some other relationship besides hasPart,
  and there may be cases where a hasPart relationship is expressed over multiple
  sentences. This measure is thus only approximate.}.

By one measure, all but one of our resources have high salience: WordNet and
ConceptNet were either hand-built, thus, by our definition, all the entries are
salient as someone thought to mention the relation. The remainder, bar PWKB,
contain entries extracted from at least one sentence, thus again by
definition, someone wrote down the relationship. The one
exception is PWKB, where the large majority (over 90\%) of the contents
were inferred through inheritance and transitivity of the hasPart
relationship, rather than directly extracted. To assess salience in PWKB, we queried
a large corpus (using the 1.7B Waterloo corpus, described earlier in Section~\ref{step1})
for sentences mentioning both entities in a PWKB hasPart relationship, for a
random sample of 1000 entries. We found that only 7.2\% pass this
``salience'' test. This suggests that PWKB, although large, contains mainly
obscure relationships. The results from the first 10 are shown in Figure~\ref{pwkb-salience}
to illustrate this.

  \begin{figure}
\centering
{\small
\begin{tabular}{|ll|} \hline
hasPart(chiropodist,liver) & hasPart(recruit,musculus sphincter ani internus) \\
hasPart(nuncio,hip)  & hasPart(Paiute,lacrimal bone) \\
{\bf hasPart(kapok,bark)} & hasPart(herring,nerve fiber) \\
hasPart(Kansa,uveoscleral pathway) & hasPart(Kickapoo,Golgi`s cell) \\
hasPart(krigia,section) & hasPart(ayatollah,musculus adductor magnus) \\ \hline
\end{tabular}
}
\caption{A random selection of 10 entries in PWKB. Only one of these (bold)
  has a Waterloo sentence co-mentioning the entities (here: ``Tree bark from the Kapok tree.''),
  loosely indicating the low salience (obscurity) of PWKB's contents.}
\label{pwkb-salience}
  \end{figure}

\eat{
  \begin{figure}
\centerline{
 \fbox{%
   \parbox{0.8\columnwidth}{
     {\small
hasPart(chiropodist,liver): (no sentence found) \\
hasPart(nuncio,hip): (no sentence found) \\
{\bf hasPart(kapok,bark): Found: "Tree bark from the Kapok tree."} \\
hasPart(Kansa,uveoscleral pathway): (no sentence found) \\
hasPart(krigia,section): (no sentence found) \\
hasPart(recruit,musculus sphincter ani internus): (no sentence found) \\
hasPart(Paiute,lacrimal bone): (no sentence found) \\
hasPart(herring,nerve fiber): (no sentence found) \\
hasPart(Kickapoo,Golgi`s cell): (no sentence found) \\
hasPart(ayatollah,musculus adductor magnus): (no sentence found)
}}}}
\caption{A random selection of 10 entries in PWKB, plus a retrieved sentence containing
  both entities (if present) in the Waterloo corpus. We use the existence of 
  a retrieved sentence as a weak indicator of the salience of the hasPart relationship.}
\label{pwkb-salience}
  \end{figure}
}

As a more systematic measure of salience, again approximated as there being
a sentence somewhere co-mentioning the entity pairs, we search the Web.
We do this using the Bing search engine, using the entity pair in a hasPart relation
as the search query, and then searching the snippets in the first page of search results
for their co-mention in a sentence. This is necessarily approximate for several reasons:
the first page of search results may miss a co-mentioning sentence elsewhere on
the Web; if a co-mention is found, it may not be expressing a hasPart relation (or may
not even be a grammatical sentence); the co-mention may be using the words in a
different sense; etc. Nevertheless, as these limitations apply to all resource
tested, the comparison is still valid, if only as an approximation.

We randomly selected 1000 hasPart relations from each of our resources,\footnote{
except for TupleKB, which only contains a total of 898 hasPart relations (we use them all).}
and ran this test. The results are shown in Table~\ref{salience-table}. 
Although necessarily approximate, these indicate that all the resources bar Webchild and PWKB have
high salience. (They also indicate the presence of a few obscure hasPart relationships in
the manually created resources, e.g., hasPart(``dactyl'',''nail'') in WordNet,
hasPart(``afterdamp'',''dihydrogen sulfide'') in ConceptNet). 

\begin{table}
  \centering
  {\small
  \begin{tabular}{|l|c|} \hline
          {\bf Source} & {\bf Salience (\%)} \\ \hline
          Wordnet 3.0 & 73.4 \\
          ConceptNet 5.6.0 & 86.5 \\
          Quasimodo 1.2  & 82.3 \\
          TupleKB & 95.4 \\
          Webchild & 18.4 \\
          PWKB & 12.1 \\
          hasPartKB (ours) & 79.2 \\ \hline
  \end{tabular} 
}
  \caption{Salience of different resources. Here, we consider a hasPart relation as salient if it is stated somewhere on the Web.
    As an approximation of this, we search for a Web sentence co-mentioning both entities in the hasPart relation, and
    report the percent of hasPart relations where this search is successful in the first page of results.}
\label{salience-table}
\end{table}

\vspace{-2mm}
\subsection{Same-Corpus Comparison \label{tuplekb-comparison}}
\vspace{-2mm}
As we have used the same source corpus as for the TupleKB (the Waterloo corpus, Section~\ref{step1}),
we have the unusual opportunity to directly compare the different extraction
techniques used, given the same input. Most importantly,
we observe hasPartKB has both a significantly higher overall yield of
hasPart relations (Table~\ref{yields}) at higher precision (Table~\ref{precision}),
given the {\it same input corpus}. Although the TupleKB was targetting a wide
variety of relations, rather than just hasPart, this provides
an indication that our use of generic sentences, rather than
an extraction pipeline over all sentences, has yielded an
advantage, at least for hasPart extraction.

\vspace{-2mm}
\section{Limitations and Discussion}
\vspace{-2mm}
Our hasPartKB complements existing resources, and is the first one that is all three
of: accurate ($\approx$90\%), high coverage of common terms (5th Grade vocabulary);
and salient (covers relationships a person may mention).
However, the extractor still makes errors. From a random sample of
incorrect extractions, we identify the following error categories:

\begin{des}
\item[{\bf 1. Ambiguous Relations ($\approx$35\% of cases):}]
In some cases the linguistic expression of hasPart is ambiguous,
e.g., ``of'' can denote multiple relations, not just meronymy.
Ideally, the trained model will correctly distinguish when
hasPart is intended, but in practice errors occur. For example, from:
\begin{myquote}
  {\it ``Inflammatory cells consist of lymphoid cells, as well as mast cells, ...''}
\end{myquote}
we incorrectly extract {\it hasPart(``inflammatory cells'',''lymphoid cells'')}. Here,
the model has taken ``consist'' to indicate meronymy.

\item[{\bf 2. Incorrect Pairing ($\approx$30\%):}]
Sometimes the model incorrectly identifies a hasPart relationship between two distant spans, for example, from:
\begin{myquote}
{\it Slugs belong to families which include snails with shells.}
\end{myquote}
we incorrectly extract {\it hasPart(``family'',''shell'')}. Similarly, from:
\begin{myquote}
  {\it Soil contains nutrients that plants feed on through their roots.} 
\end{myquote}
we incorrectly extract {\it hasPart(``soil'',''root'')}. Again, additional training data may help alleviate such mistakes.

\item[{\bf 3. Contextual Relationships ($\approx$20\%):}]
For $\approx$25\% of the errors, an over-general, contextual term was extracted, for
example, from:
\begin{myquote}
  {\it Most species have specialized breathing siphons.}
\end{myquote}
the extractor finds the over-general {\it hasPart(``species'',''breathing siphon'')}.
In this context, ``species'' is not referring all species, but species of
an organism mentioned in the previous sentence. 
Even without cross-sentence
contextualization, such errors can arise, e.g., from 
{\it Birds are animals with beaks and feathers.} we extract {\it hasPart(``animal'',``beak'')},
an over-general extraction.

\item[{\bf 4. Metonymy and Factual Errors ($\approx$10\%):}] 
In some cases, the original sentence is incorrect from a literal reading,
either due to a factual error or (more commonly) metonymy \cite{fass1997processing}. For example:
\begin{myquote}
  {\it Spider monkeys have no thumbs, so their grasping is with four fingers.''}
\end{myquote}
We incorrectly extract {\it hasPart(``spider monkey'',''four finger'')}.
In fact, the sentence is metonymically referrring to the (unstated) {\it hand}
of a spider monkey as having four fingers. Similarly:
\begin{myquote}
  {\it "All ages have strong jaws with a hooked beak, strong claws, and a long saw-toothed tail."}
\end{myquote}
produces {\it hasPart(``age'',''beak,claw'')}. Here the writer used the phrase
``All ages'' to metonymically refer to ``Young snapping turtles of all ages''.

\item[{\bf 5. Metaphor and Other Errors ($\approx$5\%):}] As an example of metaphor, from:
\begin{myquote}
{\it ``Chinchillas are creatures of habit with strong internal clocks.''}
\end{myquote}
we extract {\it hasPart(``creature'',''clock'')}, whereas here ``clock'' is meant metaphorically
rather than literally (and should have been paired with ``chinchilla'' not ``creature''). Similarly,
syntactic errors can occur, e.g.,
\begin{myquote}
{\it ``Defoliated trees grow replacement leaves that are high in tannin.''}
\end{myquote}
produces {\it hasPart(``replacement leave'',...)} rather than {\it hasPart(``replacement leaf''...)}.

\end{des}

\eat{
  \vspace{-2mm}
  \subsection{The Semantics of hasPart at the Boundaries}
\vspace{-2mm}
As well as the specific error categories, above, we note that the semantics
of ``having a part'' itself has some ambiguity for boundary cases along three dimensions:

\begin{des}
\item[(a)] What exactly constitutes a part: For example, is wallpaper {\it part of} a
room or {\it contained in} the room?
\item[(b)] What exactly constitutes the entities being related: For example, does a
person's hand have four fingers or five?
\item[(c)] What quantification is sufficient: For example, does an airplane have
  a propeller?
\end{des}

\noindent
Although these issues only affect boundary cases, they are important to note
for future development.
}
\vspace{-2mm}
\section{Conclusion}
\vspace{-2mm}
Meronymic relations are one of the most important relationships between
entities. To complement existing resources, we have presented
a new knowledge-base of hasPart relationships, constructed
in a novel way by using generic sentences as a source of
knowledge. Empirically, the approach has yielded the first resource that is all
three of: accurate ($\approx$90\%), salient (covers relationships
a person may mention), and has high coverage of common terms.
In addition, it contains information about quantifiers, argument modifiers,
and links the entities to appropriate concepts in Wikipedia and WordNet.
The KB is available for the community at \kburl
 
\bibliography{references}
\bibliographystyle{plainnat}

\clearpage
\section*{Appendix: Word Sense Disambiguation Model}

Our word sense disambiguation model is a variant of the GlossBERT model of \citet{Huang2019GlossBERTBF}, which uses synset glosses in WordNet to disambiguate words. More specifically, given a sentence $s$ and a set of target words $\{ w_{1},w_{2},...,w_{n}\}$ in that sentence, the BERT encoder is used to score each $w_{j}$ against a set of candidate WordNet glosses $\{g^{(w_{j})}_{1},g^{(w_{j})}_{2},..., g^{(w_{j})}_{m}\}$ using the following format:  $$\texttt{$s_{w_{j}}^{g_{o}}$ := [CLS] $s$ [SEP] $w_{j}$: $g^{(w_{j})}_{o}$ [SEP]}.$$ This model is trained on the SEMCOR 3.0 training corpus \cite{raganato2017word} using a standard sentence-pair classification loss over the hidden state of the classifier token  for (denoted as $\textbf{c}_{s,w_{j}}^{g_{o}} = \textbf{BERT}(s_{w_{j}}^{g_{o}}) \in \mathbf{R}^{512}$)  and an additional classification layer (\texttt{c}$(\cdot)$).


In our version, the more recent RoBERTa encoder is used in place of BERT. We also employ an alternative multiple-choice ranking loss and format, following that used in \cite{Richardson2019WhatDM} where the loss of the correct gloss $g^{*}$ is computed over all other glosses (i.e., given $\textbf{c}_{s,w_{j}}^{g^{*}}$ and the probability $p_{w,s}^{g^{*}} \propto e^{\texttt{c}(\textbf{c}_{s,w_{j}}^{g^{*}})}$ of $g^{*}$ over all alternative glosses, our loss over our training dataset $D$ is given as $\mathcal{L} = \sum_{(w,s) \in D} -\log p_{w,s}^{g^{*}}$).

Results on a standard WSD benchmark suite \cite{Navigli2017WordSD}, comparing with two state-of-the-art BERT-based WSD algorithms, are shown in Table~A1. This suggests that our model has comparable (even a little better) performance than these earlier models.

\begin{table}[h]
\centering
{\small
\setlength\tabcolsep{3pt}    	
\begin{tabular}{|l|lllll|} \hline
          & {\bf Semeval'07} & {\bf Senseval 2} & {\bf Senseval 3}  & {\bf Semeval'13} & {\bf Semeval'15} \\
{\bf Models} & {\bf (dev)} & {\bf (test)} & {\bf (test)} & {\bf (test)} & {\bf (test)} \\ \hline
GlossBERT & 72.5 & 77.7 & 75.2 & 76.1 & 80.4 \\

BERT-WSD  & --   & 76.4 & 74.9 & 76.3 & 78.3 \\
{\bf RoBERTa-WSD (ours)}                    & 74.9 & 80.2 & 77.2 &  79.0 & 82.3 \\ \hline
\end{tabular}
} 
\caption*{Table A1: Performance of our RoBERTa-WSD Algorithm (Section~\ref{wsd}) on the standard WSD benchmark suite \cite{Navigli2017WordSD}
against two state-of-the-art algorithms,
GlossBERT \cite{Huang2019GlossBERTBF} and BERT-WSD \cite{Du2019UsingBF}.}
\end{table}





\eat{
\section*{A2. Reference Parts Lists for Coverage Case Study}

The following are the independently authored parts lists used in the coverage case-study (Section~\ref{case-study}):
\begin{ite}
\item {\bf Dog}: eye, cheek, tongue, neck, shoulder, chest, elbow, forelimb, wrist, claw, paw, toes, hock, tail, abdomen(belly), thigh, wither, back, nose, muzzle, ear, mouth, whiskers
\item {\bf Ribosome}: small subunit, large subunit, A site, P site, E site
\item {\bf Snail}: eyespots, foot, shell, head, mouth, respiratory pore, tentacles
\item {\bf Maple}: leaf, flower, seeds, twig, trunk, bark, root, branch, crown
\item {\bf Hippo}: toes, eyes, nostrils, skin, legs, tail
\item {\bf Fish}: caudal fin, anal fin, pelvic fin, pectoral fin, gill, gill cover, mouth, eye, nostril, dorsal fins, scales, lateral line
\end{ite}
}

\end{document}